# New design of smooth PSO-IPF navigator with kinematic constraints

**Mahsa Mohaghegh[1,2], Hedieh Jafarpourdavatgar[3], Samaneh Alsadat Saeedinia[4]**

[1] School of Engineering, Computing and Mathematical Sciences, Auckland University of Technology (AUT), Auckland, New Zealand
[2] Faculty of Design and Creative Technologies, AUT, Auckland, New Zealand
[3] Control Engineering Department, Amirkabir University, Tehran, Iran
[4] Complex System Laboratory, Iran University of Science and Technology, Tehran, Iran
Corresponding author: Mahsa Mohaghegh (e-mail: mahsa.mohaghegh@aut.ac.nz).

**ABSTRACT** Robotic applications across industries demand advanced navigation for safe and smooth movement. Smooth path planning is crucial for mobile robots to ensure stable and efficient navigation, as it minimizes jerky movements and enhances overall performance Achieving this requires smooth collision-free paths. Partial Swarm Optimization (PSO) and Potential Field (PF) are notable path-planning techniques, however, they may struggle to produce smooth paths due to their inherent algorithms, potentially leading to suboptimal robot motion and increased energy consumption. In addition, while PSO efficiently explores solution spaces, it generates long paths and has limited global search. On the contrary, PF methods offer concise paths but struggle with distant targets or obstacles. To address this, we propose Smoothed Partial Swarm Optimization with Improved Potential Field (SPSO-IPF), combining both approaches and it is capable of generating a smooth and safe path. Our research demonstrates SPSO-IPF's superiority, proving its effectiveness in static and dynamic environments compared to a mere PSO or a mere PF approach.

## I. INTRODUCTION

Robotic navigation in dynamic and uncertain environments poses formidable challenges, demanding the development of efficient and adaptable algorithms. Optimization algorithms are indispensable in mobile robot navigation, facilitating the discovery of optimal solutions to intricate problems. These algorithms empower robots to meticulously plan their paths, accurately localize themselves, and dynamically avoid obstacles in real time, thereby significantly enhancing navigation performance [1]. Among the plethora of optimization methodologies, Particle Swarm Optimization (PSO) and Potential Fields (PF) stand out as widely utilized approaches with extensive applications in mobile robot navigation [2]. Particularly beneficial in scenarios with complex constraints or where energy consumption minimization is paramount, PSO and PF algorithms excel in addressing the multifaceted demands of navigation tasks.

Numerous studies and articles delve into the application of these optimization algorithms in mobile robot navigation, underscoring their pivotal role in path planning, environmental adaptability, and navigation efficiency [3][4][5]. Within these algorithms, particles traverse the search space to ascertain the optimal path, typically defined by a fitness function. However, despite its efficacy in finding optimal paths, PSO often grapples with generating smooth trajectories [4]. This lack of smoothness can result in jerky or discontinuous motions, compromising path traversal efficiency and potentially destabilizing the system.

In contrast, PF emerges as a probabilistic filtering technique adept at estimating the state of dynamic systems based on noisy sensor measurements [5]. In PF path planning, a collection of particles representing feasible paths propagates throughout the environment. Each particle's weight is determined by its likelihood given the sensor measurements, with resampling procedures implemented to manage particle distribution. Yet, achieving path smoothness remains a challenge for PF, especially in scenarios where the robot veers significantly from the target, potentially leading to collision risks due to excessive attraction toward obstacles [6]. Additionally, situations may arise where obstacle repulsion and target attraction forces nullify each other, trapping the robot indefinitely [7]. To alleviate these concerns, researchers have proposed modifications to the artificial potential field approach, introducing novel sets of potential fields [8].

Moreover, the resampling process in PF can inadvertently introduce variability and diminish path smoothness [5].

Robotic navigation in dynamic and uncertain environments presents a myriad of challenges, notably in achieving smooth path planning with PSO and PF. One fundamental hurdle lies in balancing exploration and exploitation [6]. While exploration seeks new and potentially better paths, the dominance of this factor can yield erratic and disorganized paths. Conversely, an overemphasis on exploitation may trap the search in local optima, resulting in suboptimal trajectories. Thus, meticulous tuning of this trade-off is pivotal for both path efficiency and smoothness.

Furthermore, the discrete representation of paths in PSO and PF, often grid-based, introduces artifacts and discontinuities [7]. Researchers propose solutions like higher-resolution grids or alternative spatial representations such as curves or splines to mitigate this challenge [7][8]. However, these approaches may introduce additional computational complexity and trade-offs between smoothness, accuracy, and real-time performance.

Moreover, the complexity and dynamic nature of environments significantly impact path smoothness [9][10]. PSO and PF may struggle in complex environments with numerous obstacles or narrow passages due to limited exploration capabilities. In highly dynamic environments, rapid changes may render generated paths obsolete, necessitating abrupt corrections or replanning.

Integration of PF and PSO techniques emerges as a promising solution for optimal path planning, synergizing their strengths to enhance mobile robot navigation capabilities. This integration offers potential solutions to address the challenges of exploration-exploitation balance, path representation, and environmental dynamics, ultimately facilitating smoother and more efficient navigation [11].

Our contribution lies in pioneering the fusion of an Improved Particle Filter (IPF) with Particle Swarm Optimization (PSO) to achieve unprecedented smoothness in path planning. By enhancing the traditional PF with a novel objective function tailored for smoothness, we introduce a paradigm shift in mobile robot navigation.

Key to our innovation is the meticulous adjustment of the objective function to prioritize smooth trajectories, a departure from conventional PF methods. Through rigorous experimentation and comparison with both PSO and standalone IPF approaches, we demonstrate the superiority of our hybrid method in generating smoother paths while maintaining efficiency and robustness.

This novel integration addresses a critical gap in existing navigation algorithms, where smoothness is often sacrificed for optimization objectives by considering robots' kinetic constraints. Our method not only outperforms conventional PSO and IPF solitary approaches but also offers a versatile solution adaptable to diverse environments and constraints.

Overall, our work presents a groundbreaking advancement in mobile robot navigation, ushering in a new era of smoother, more efficient path planning. Through the integration of Improved PF with PSO and the optimization of our objective function, we pave the way for enhanced navigation capabilities in dynamic and uncertain environments.

The paper is organized into several sections to systematically present the research findings. The Materials and Methods section details the proposed approach of combining Smooth Particle Swarm Optimization (SPSO) with Improved Potential Fields (IPF) to enhance path planning smoothness in mobile robots. This section outlines the modifications made to the objective function to prioritize smoothness and describes the experimental setup and parameters used for simulation. The Results section presents the simulation outcomes, showcasing the effectiveness of the SPSO-IPF approach in generating smoother paths compared to traditional PSO and IPF solitary methods. Simulation results are illustrated and discussed comprehensively, highlighting the advantages of the proposed approach in dynamic and uncertain environments.

## II. MATERIAL AND METHODS

In this section, we present the methodology employed for path planning, focusing on a novel approach integrating Smooth Particle Swarm Optimization (PSO) with an Improved Potential Fields (IPF) method. This approach addresses the challenges of path planning for mobile robots while considering kinematic constraints. The integration of PSO and IPF aims to generate smooth and collision-free paths in dynamic environments, ensuring optimal trajectory lengths.

The smooth PSO-IPF approach results from advancements in optimization techniques, particularly suited for geometric path planning in autonomous and robotics applications. By incorporating kinematic constraints, the proposed approach seeks to improve the efficiency and safety of mobile robot navigation.

The following sections outline the key components of the smooth PSO-IPF approach, detailing its formulation and implementation to achieve robust and effective path-planning solutions under dynamic conditions.

### A. HYPOTHESES

To design a path planning algorithm, it's crucial to identify the problem priorities and constraints. In this study, the following problem priorities are considered:

- Ensure collision-free movement from start to goal in a dynamic environment.
- Ensure smooth traversal along the path.
- Consider near-optimal path length.

The constraints for the problem are defined as follows:

- Linear velocity of the robot is restricted to be lower than 0.8 m/s in x-y coordinates.
- Angular velocity is limited to $\frac{\pi}{6}$ rad/s.

While achieving an optimal path length is a priority, global approaches are favored, acknowledging their limitations in ensuring safety in dynamic environments. Therefore, a hybrid approach combining reactive and classic methods is proposed. Specifically, the Potential Fields (PF) method is integrated with the Particle Swarm Optimization (PSO) algorithm to address the identified priorities within the mentioned constraints. The total potential field serves as the fitness function, optimized by PSO. Additionally, modifications are made to the cell decomposition to incorporate angular and linear displacement constraints, with kinematic constraints also considered within the IPF framework. The problem definition and methodology are outlined in the accompanying flowchart, Fig.1.

### B. Designing SPSO_IPF method

Let's define a hybrid path planning problem, step by step according to the given flowchart. Total Potential Field $U(q)$ is a superposition of attractive potential $U_{att}(q)$ and repulsive potential $U_{rep}(q)$, described by:

$$U(q) = U_{att}(q) + U_{rep}(q) \qquad (1)$$

Where $q$ is the position *[x y]* in the global coordination. Conventional PF methods fail in situations where the target is far away, or when the robot is located very close to the obstacles, and when the calculated potential field is zero. To address these issues, we use an improved version of the PF method in which the attractive and repulsive potentials are modified and described as follows [8]:

$$U_{att}(q) = \begin{cases} \dfrac{1}{2}\varepsilon.d^2(q,q_{Goal}) & d(q,q_{Goal}) \leq d^*_{Goal} \\ d^*_{Goal}\varepsilon.d(q,q_{Goal}) - \dfrac{1}{2}\varepsilon.\left(d^*_{Goal}\right)^2 & d(q,q_{Goal}) > d^*_{Goal} \end{cases} \qquad (2)$$

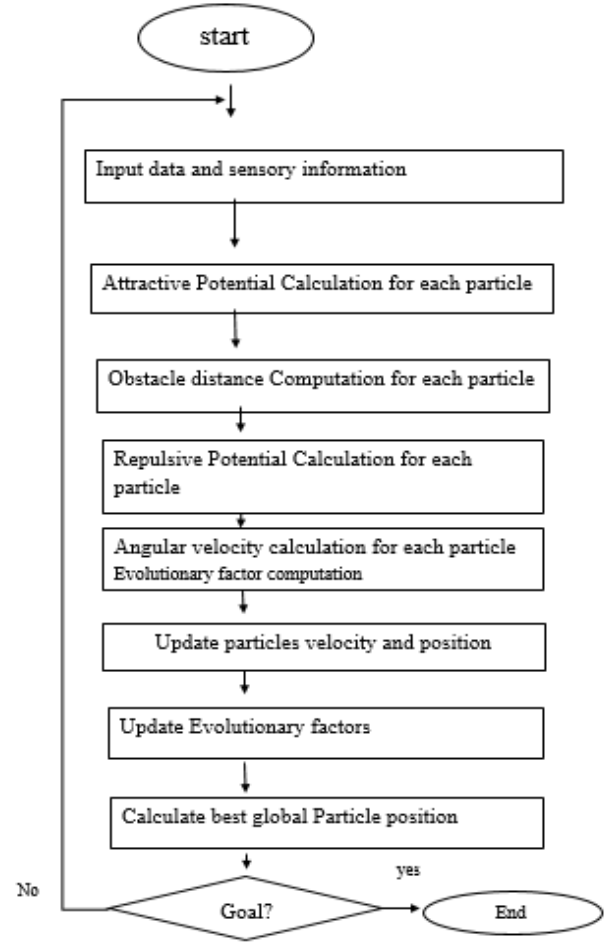

start

Input data and sensory information

Attractive Potential Calculation for each particle

Obstacle distance Computation for each particle

Repulsive Potential Calculation for each particle

Angular velocity calculation for each particle
Evolutionary factor computation

Update particles velocity and position

Update Evolutionary factors

Calculate best global Particle position

Goal? — yes → End

No

**FIGURE 1  Flow chart of path planning method of PSO integrated by PF**

where, $\varepsilon$ is the attraction gain, $d(q,q_{Goal}) = \left(\left(q - q_{Goal}\right)\left(q - q_{Goal}\right)^T\right)^{\frac{1}{2}}$ is the distance between robot and goal, $d^*_{Goal}$ is the threshold value that defines the distance between the robot and the goal which is compared for the choice between conic and quadratic potential. The attractive force which is the gradient of the function (2) is defined below:

$$F_{att}(q) = -\nabla U_{att}(q) = \begin{cases} -\varepsilon.\left(q - q_{Goal}\right) & d(q,q_{Goal}) \leq d^*_{Goal} \\ -d^*_{Goal}\varepsilon\dfrac{\left(q - q_{Goal}\right)}{d(q,q_{Goal})} & d(q,q_{Goal}) > d^*_{Goal} \end{cases} \qquad (3)$$

Correspondingly, repulsive potential can be described by:

$$U_{rep} = \begin{cases} \dfrac{1}{2}\eta \left( \dfrac{1}{d\left(q, q_{obs}\right)} - \dfrac{1}{d_0} \right)^2 d^*\left(q, q_{Goal}\right) & d(q, q_{obs}) \le d_0 \\ \\ 0 & d(q, q_{obs}) > d_0 \end{cases} \quad (4)$$

where $d\left(q, q_{obs}\right)$ is the distance between the obstacle and the robot, $d_0$ is the threshold value and n is the positive constant and $\eta$ is a positive scaling factor. The repulsive force of (5) is calculated as follows:

$$F_{rep} = -\nabla U_{rep}(q) = \begin{cases} F_{rep1} + F_{rep2} & d(q, q_{obs}) \le d_0 \\ 0 & d(q, q_{obs}) > d_0 \end{cases} \quad (5)$$

$$F_{rep1} = \eta \left( \frac{1}{d\left(q, q_{obs}\right)} - \frac{1}{d_0} \right) \frac{d^*\left(q, q_{Goal}\right)}{d^2\left(q, q_{Goal}\right)} \frac{(q - q_{obs})}{d\left(q, q_{obs}\right)} \quad (6)$$

$$F_{rep2} = -\frac{n}{2}\eta \left( \frac{1}{d\left(q, q_{obs}\right)} - \frac{1}{d_0} \right)^2 d^{*-1}\left(q, q_{Goal}\right) \frac{(q - q_{Goal})}{d\left(q, q_{Goal}\right)} \quad (7)$$

This paper suggests to adapt $d_0$ to adapt distance threshold value based on velocity of robots and obstacle, if the obstacle velocity is observable, defined as $V_{obs}(k)$, and the paper suggest a new predictive IPF collision avoidance criteria for repulsive force as follows:

$$d_0 = \begin{cases} \dot{q} + \dot{q}_{obs} + d_{01} & if \; \dot{q}_{obs} \;\; \text{is determined} \\ \\ \dot{q} + \max\{\dot{q}_{obs}\} + d_{01} & if \; \max\{\dot{q}_{obs}\} \;\;\; \text{is determined} \\ \\ 2\max\{\dot{q}\} + d_{01} & \text{otherwise} \end{cases} \quad (8)$$

We have assumed that the velocity of the robot, $\dot{q}$, can be observed and estimated either through the optimal particle velocity approximation (see (12)) or the dynamic equations of the robot. Similarly, in most environments, it is possible to determine the velocity or at least the maximum velocity of the obstacles. For the purpose of uncertainty considerations, '$d_{01}$' is a constant value.

According to the mentioned priorities and criteria of the navigation problem, minimization of the potential field (1) meets two criteria; namely, minimum possible path length and collision avoidance criterion. To meet the path smoothness criteria integrated with the Improved Potential field strategy, the objective function can be regarded as follows:

$$f(q) = U(q) + \left( \dot{\theta}_q - \omega_{max} \right)^2 + \left( \dot{q} - \dot{q}_{max} \right)^2 \quad (9)$$

Where $\dot{\theta}_q$ is the angular velocity, $\omega_{max}$, and are respectively the maximum angular velocity, and velocity of the robot, defined by kinematics constraints. This study proposes the following novel relationship between angular velocity, velocity, and movement. This equation is calculated based on considering a restriction on $arctan(q_{goal} - q)$ which is calculated as follows:

$$\dot{\theta}_q = \frac{(\dot{q}_{goal} - \dot{q}) \begin{bmatrix} 0 & -1 \\ 1 & 0 \end{bmatrix} \left(q_{goal} - q\right)^T}{d(q, q_{goal})^2 + \varepsilon_0} \quad (10)$$

Where $\varepsilon_0$ is a small positive constant which is considered with the aim of avoiding singularity condition. To solve the optimization problem, the PSO algorithm is utilized. To this aim, PSO algorithms employ several particles as the candidate solutions to the problem, flying around to search for the optimal solution in the search space. The velocity and position of an ith particle can be updated iteratively in the kth iteration as follows [14]:

$$V_i(k+1) = wV_i(k) + c_1 r_1 (q_{p_i}(k) - q(k)) + c_2 r_2 (q_g(k) - q(k)) \quad (11)$$

$$q_i(k+1) = q_i(k) + V_i(k+1) \quad (12)$$

Where $W$ represents the inertia weight. $c_1$ and $c_2$ are the coefficients of particle acceleration, mostly selected in the range of [0,2]. $r_1$ and $r_2$ denotes random uniform variables in the range [0,1], $q_{p_i}$ and $q_g$ are respectively local and global best position of particles, determined as follows.

$$q_{p_i}(i,t) = \arg\min_{k=1,...,t} f(q_i(k)), \quad i \in \left\{1,...,N_p\right\} \quad (13)$$

$$q_g(t) = \arg\min_{\substack{k=1,...,t \\ i \in \{1,...,N_p\}}} \left[ f(q_i(k)) \right], \quad (14)$$

Where $Np$ is the number of particles. The algorithm for the integration of PSO and Improved PF (PSO-IPF) is described below:

---

**Algorithm1**

---

*Step1.Initialization*
For each Particle $i=1,..., Np$, do
  a) determine Lower band $L_b$ and Upper band $U_b$ of particles' positions, based on kinematic constraints
  b) Initialize the particles' position $P_{it} = q_i(t)$, uniformly in the range of $[L_b, U_b]$.

c) Initialize $q_{p_i}(i,0) = P_{i0}$ and
$q_g(0) = \arg\min[f(P_{i0})]$

d) Initialize Velocity $v_i(0)$ uniformly, according to the kinematic constraints

*Step 2. Repeat by the time that the minimum objective value is achieved, do*

a) Pick random numbers $r_1$ and $r_2 \in Uniform([0,1])$

b) Update the Potential Field value for each particle (1)

c) Update the Velocity and position of each particle ((11) and (12))

d) Update $q_{p_i}$ and $q_g$ ((13) and (14))

e) Update time epoch $t \leftarrow t+1$

*Step 3.* $q_g$ is the best-found solution at time epoch t

## III. Results

Algorithm 1 is simulated to evaluate the performance of the designed SPSO-IPF path planning strategy. Simulation is conducted by Python 3.10. Figure 2,3, and 4 reveal several samples of simulated static environments and the generated path by the proposed SPSO-IPF algorithm.

Figures 2 to 4 demonstrate the performance of the proposed method in three environment samples.

It has been observed that the proposed algorithm may not always ensure smoothness in certain situations, particularly when the robot is in close proximity to the target. This can be considered a limitation of the method. However, the best run of the algorithm satisfies the smoothness requirement as defined in the objective function.

Figure 5 indicates several examples of executions of the SPSO-IFP method in three different obstacle arrangements. Path smoothness in all executions is almost detectable. The PSO algorithm inherently is categorized into the search–based. and sequence–based techniques. Therefore, several iterations are required to find the near-optimal path. $c_1$ and $c_2$ are considered 2 in each iteration. Figure 5 indicates a dynamic example of the environment.

Figure 5 depicts the path generation performance by the designed SPSO-IPF method. Since, for real-time applications, one run per epoch is considered, only the collision avoidance criterion is seen in the generated path, smoothness and shortest length path criteria do not significantly play their roles in the generated way.

As shown in Figure 6, the designed SPSO-IPF method improves path generation by reducing path length and increasing smoothness, compared to a simple PSO.

Figure 7 illustrates SPSO_IPF performance in a complex environment, consisting of both static and dynamic obstacles and it compares with the non-Smooth PSO_IPF approach [15].

Figure 7 illustrates how the proposed method can outperform non-smooth PSO_PF [15], both in terms of path length and generating a smooth path within a complex environment.

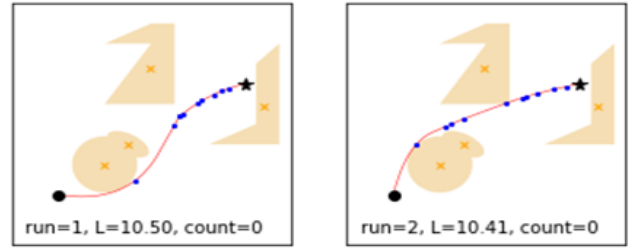

**FIGURE 2** Two runs of the SPSO-IPF method in the first environment sample, run2 with lower length path L=10.41 is the best solution.

## IV. Conclusion

This paper presents an innovative approach known as path-smoothed PSO_IPF, which not only incorporates kinematic constraints but also enhances collision avoidance criteria within the Potential field objective function. It underscores the significance of taking into account various factors, such as obstacle avoidance, robot speed, and environmental dynamics, when formulating optimal path-planning strategies. While emphasizing the criticality of path smoothness, particularly in static environments, the study acknowledges its potential decline in importance in dynamic settings near the target. Nonetheless, the proposed technique demonstrates promising outcomes by generating secure, concise, and notably smooth paths in both static and dynamic scenarios. Moreover, this exploration gives important experiences into the improvement of productive navigation systems for autonomous robots. By prioritizing problem constraints and adopting a hybrid approach that combines reactive and classical techniques, it offers a framework for creating effective path-planning solutions. This approach addresses the multifaceted challenges of navigation, paving the way for enhancing the autonomy and adaptability of robotic systems in diverse real-world environments. In conclusion, the findings of this study have practical implications for the implementation of autonomous navigation systems. By integrating cutting-edge methodologies, such as PSO, and improving PF while considering kinematic constraints, this research sets the stage for the development of robust and adaptable robotic platforms capable of navigating complex environments with precision and efficiency.

### ACKNOWLEDGMENT
The authors express their gratitude and appreciation to Auckland University of Science and Technology (AUT) and its sponsors for their invaluable support and contribution to our research. We are grateful for the opportunity to conduct our research with the generous sponsorship provided by AUT, which has been instrumental in facilitating our project's success.

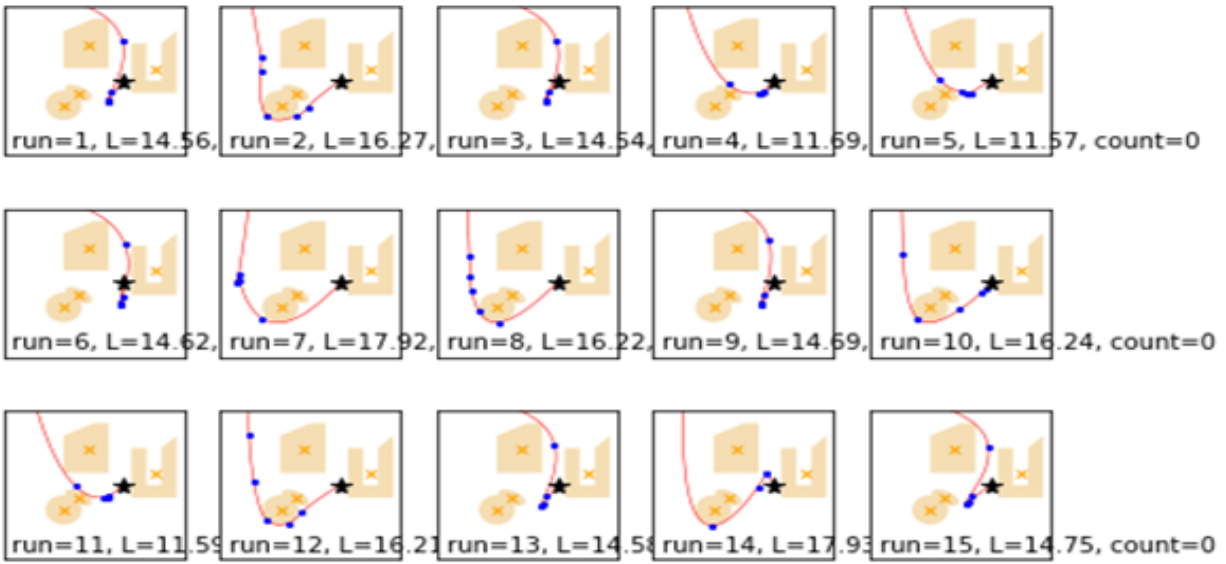

**FIGURE 4** Fifteen runs of SPSO-IPF in the second environment sample with different start and goal position, the best solution is run=5 with length path L=11.57.

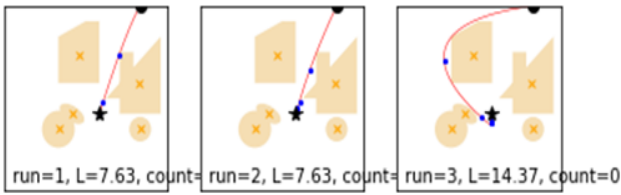

**FIGURE 3** Three runs of SPSO_IPF in the third environment sample with different start and target positions, the best solution is run=2 with length path L=7.63

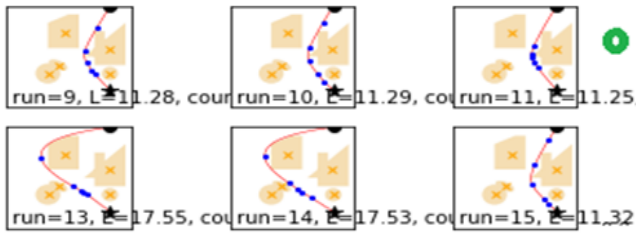

a) Proposed PSO-IPF

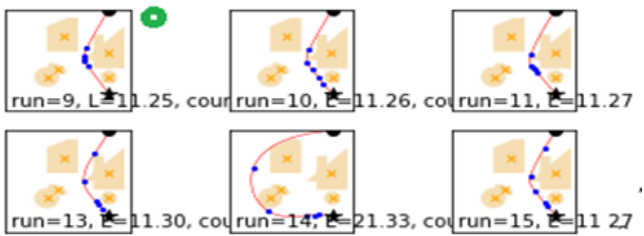

b) PSO

**FIGURE 6.** Comparing both PSO and Smoothed PSO-IPF methods

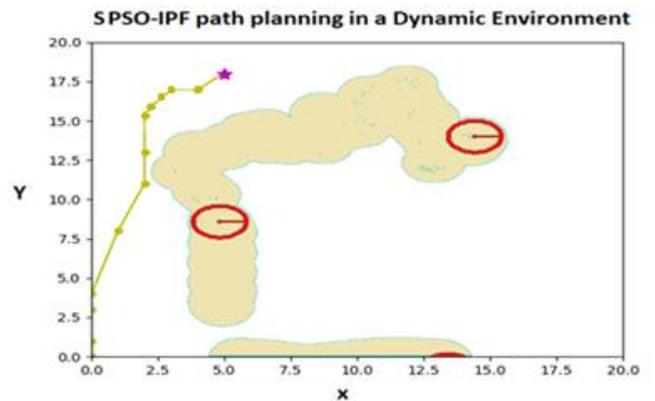

**FIGURE 5** Performance of the SPSO-IPF algorithm in a dynamic environment. A red circle shows the obstacle, the passed path by the obstacles is shown by a cream tube, a yellow line depicts the generated path and the star is the target position.

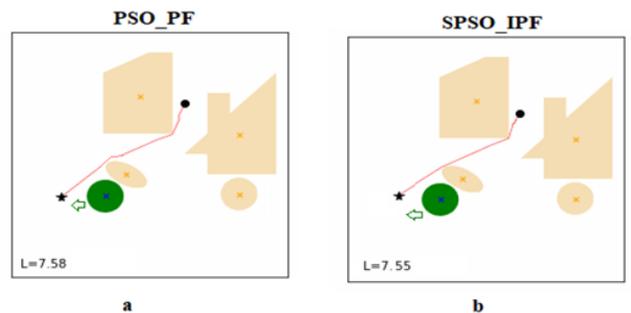

**FIGURE 7.** Path planning comparison of two methods: a) non–smooth PSO_PF [15], and b) proposed SPSO_IPF methods. The green circle indicates the moving obstacles which are directed toward the green vectors, while the other obstacles are statics

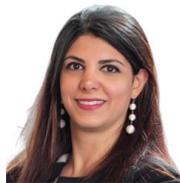

**Mahsa Mohaghegh** is an Iranian-born New Zealand computer engineer specializing in artificial intelligence and natural language processing. She is a Senior Lecturer and Director of Women in Tech at Auckland University of Technology's School of Engineering, Computer and Mathematical Sciences. Mohaghegh completed her Bachelor's degree in computer engineering and her Master's degree in computer architecture before earning her doctorate in computer engineering from Massey University in 2013. She has been involved with Google's Computer Science for High Schools program since 2013 and runs workshops in Auckland. Mohaghegh founded She Sharp, a women's networking group aimed at encouraging girls and young women to engage with digital industries. She has received several awards, including the Emerging Leader category in the 2013 Westpac Women of Influence Awards, the 2019 YWCA Equal Pay Champion Award, and the 2020 Massey University Distinguished Alumni Award. Mohaghegh is a well-recognized leader in AI and machine learning and is committed to promoting diversity and inclusion in the tech industry

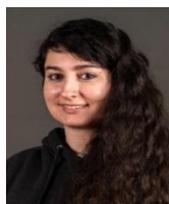

**Hedieh Jafarpourdavatgar,** born on April 16, 1989, in Iran, is a dedicated engineer with a strong interest in optimization and control systems. She earned her Bachelor's degree in Electronic Engineering from the Islamic Azad University and completed her Master's degree in Control Engineering at Amirkabir University of Tehran (Polytechnic of Tehran). Throughout her academic and professional journey, Hedieh has consistently demonstrated her commitment to tackling complex challenges in the field of control engineering, contributing to advancements in the industry.

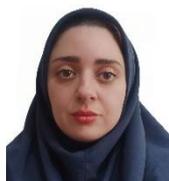

**Samaneh Alsadat Saeedinia,** born on March 29, 1989, in Qazvin, Iran, is a control engineer, holding a Bachelor of Science degree from Imam Khomeini International University and a Master of Science degree from Iran University of Science and Technology (IUST) (GPA: A). With a strong academic foundation, she is currently on the cusp of completing her Ph.D. in control engineering at IUST. Samane-Alsadat's research expertise encompasses a wide range of applications in control theory and AI, demonstrated through her roles as a Research Assistant and Project Manager